%% file: main.tex
\pdfoutput=1             

\documentclass[preprint,12pt,authoryear]{elsarticle}

\makeatletter
\def\ps@pprintTitle{%
   \let\@oddhead\@empty
   \let\@evenhead\@empty
   \let\@oddfoot\@empty
   \let\@evenfoot\@oddfoot}
\makeatother

\usepackage{setspace} 
\usepackage{lineno} 
\usepackage[nohyperlinks]{acronym}
\usepackage[hidelinks]{hyperref}
\usepackage{wrapfig} 
\usepackage{lineno} 
\usepackage{appendix}
\usepackage{tikz}
\usetikzlibrary{shapes.geometric, arrows} 
\usepackage{pdfpages}
\usepackage[T1]{fontenc}
\usepackage[utf8]{inputenc}
\usepackage{lmodern}
\usepackage{amsmath,amssymb,mathtools}
\usepackage[capitalise, noabbrev]{cleveref} 
\usepackage{graphicx}
\usepackage{booktabs}
\usepackage{siunitx}
\usepackage{doi}
\usepackage{listings}
\usepackage{caption}

\usepackage{subcaption} 
\usepackage{natbib}
\renewcommand{\citep}[1]{(\citeyear{#1})}  
\begin{document}

\doublespacing
\acresetall
\title{Data Fusion and Machine Learning for Ship Fuel Consumption Modelling - A Case of Bulk Carrier Vessel}
\author[label1,label2]{Abdella Mohamed\corref{mycorrespondingauthor}}
\ead{mohamed.abdella@tum.de}

\author[label1]{Xiangyu Hu}
\ead{xiangyu.hu@tum.de}

\author[label2]{Christian Hendricks}
\ead{christian.hendricks@everllence.com}

\cortext[mycorrespondingauthor]{Corresponding author}
\affiliation[label1]{organization={Technical Univerisity of Munich},
         addressline={Boltzmannstraße 15},
         postcode={85748},
         city={Garching},
         country={Germany}}
\affiliation[label2]{organization={Everllence (formerly: MAN Energy Solutions)},
         addressline={Stadtbachstraße 1},
         postcode={86153},
         city={Augsburg},
         country={Germany}}

\input{abstract}
\input{acronyms}
\input{introduction}
\input{literature_analysis}
\input{methodological_approach}
\input{results}
\input{conclusion}
\section*{CRediT authorship contribution statement}
\textbf{Abdella Mohamed:} Conceptualization, Methodology, Coding, Data curation, Writing, Visualization, Investigation, Validation. 
\textbf{Xiangyu Hu:} Supervision, Conceptualization, Methodology, Manuscript review and editing. 
\textbf{Christian Hendricks:} Mentorship, Conceptualization, Methodology, Manuscript review and editing. 
\section*{Acknowledgements}
This work was supported by Technical University of Munich and Everllence (formerly: MAN Energy Solutions).
The following tools have been used to produce this manuscript:
\begin{itemize}
    \item Macbook Pro M3 Pro, 36 GB RAM
    \item Draw io  
    \item Python and its Ecosystem
    \item Jupyter Lab and its Ecosystem
    \item VScode and its Ecosystem
    \item Obsidian and its Ecosystem 
    \item Latex  and its Ecosystem
    \item Zotero and its Ecosystem
\end{itemize}
\section*{Data and Code Availability}
All code and data are publically accessible. 
\listoftables
\listoffigures
\lstlistoflistings
\clearpage
\bibliography{main}
\newpage
\appendix
\setcounter{table}{0}
\section{Raw Parameters Description}\label{app:raw_parameters_description}
    \input{voyage_reports_raw_parameters_and_unit_of_measure}
    \clearpage
    \input{cmems_raw_parameters_and_unit_of_measure}
    \input{era5_raw_parameters_and_unit_of_measure}
\section{Data Distribution}\label{app:data_distribution}
\begin{figure}
    \begin{subfigure}[b]{0.45\textwidth}
        \centering
        \includegraphics[width=\textwidth]{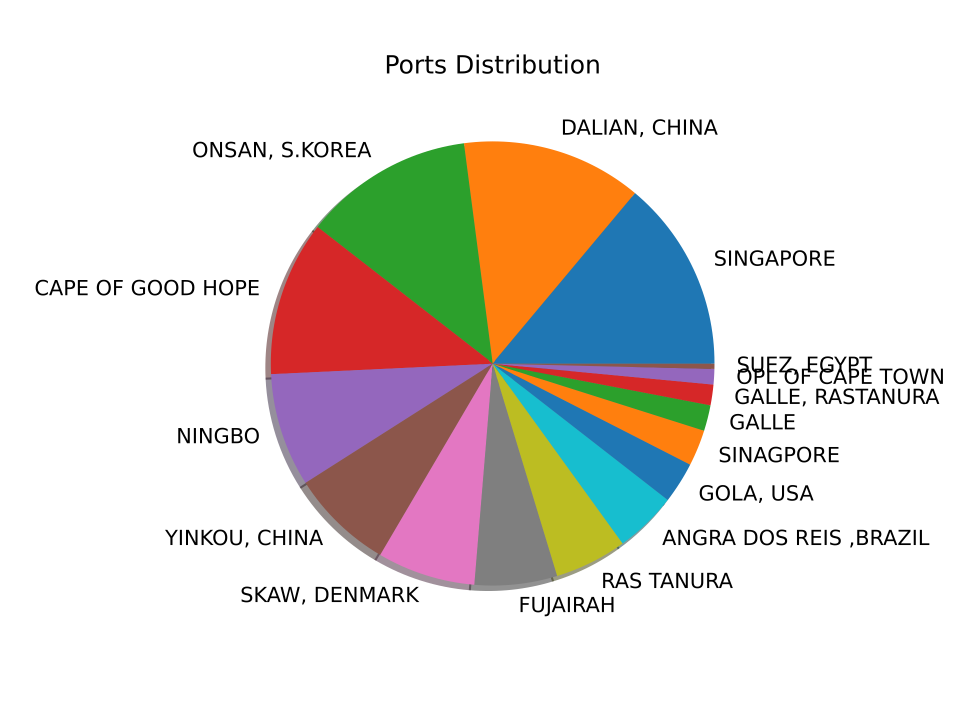}
        \caption{Ports Distribution}
        \label{fig:ports_disctribution}
    \end{subfigure}
    \begin{subfigure}[b]{0.45\textwidth}
        \centering
        \includegraphics[width=\textwidth]{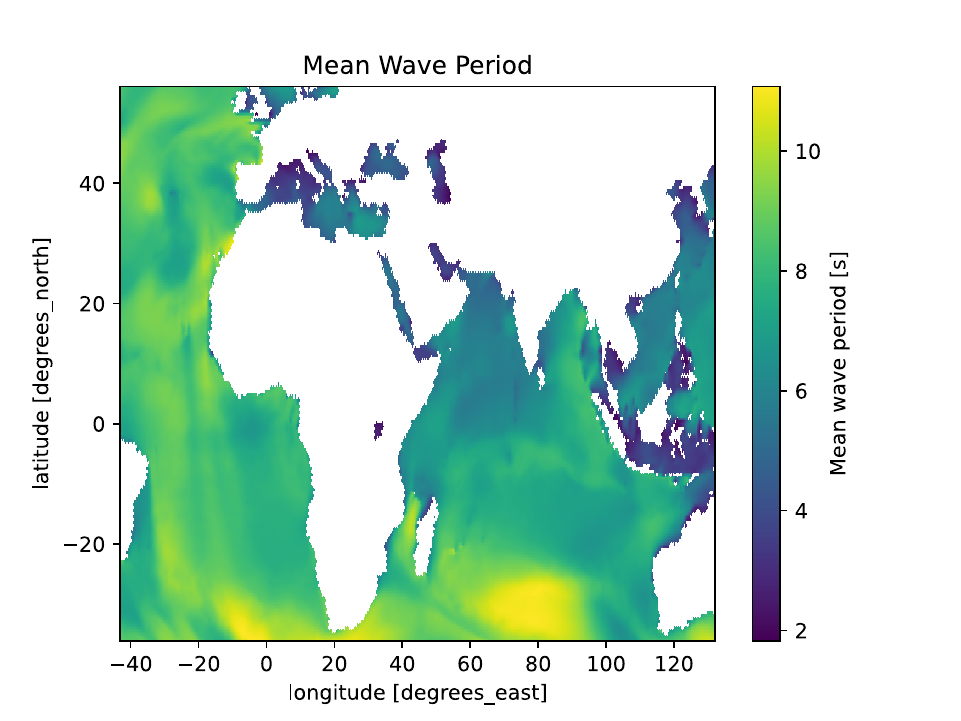}
        \caption{Mean Wave Height}
        \label{fig:mean_wave_height}
    \end{subfigure}
    \begin{subfigure}[b]{0.45\textwidth}
        \centering
        \includegraphics[width=\textwidth]{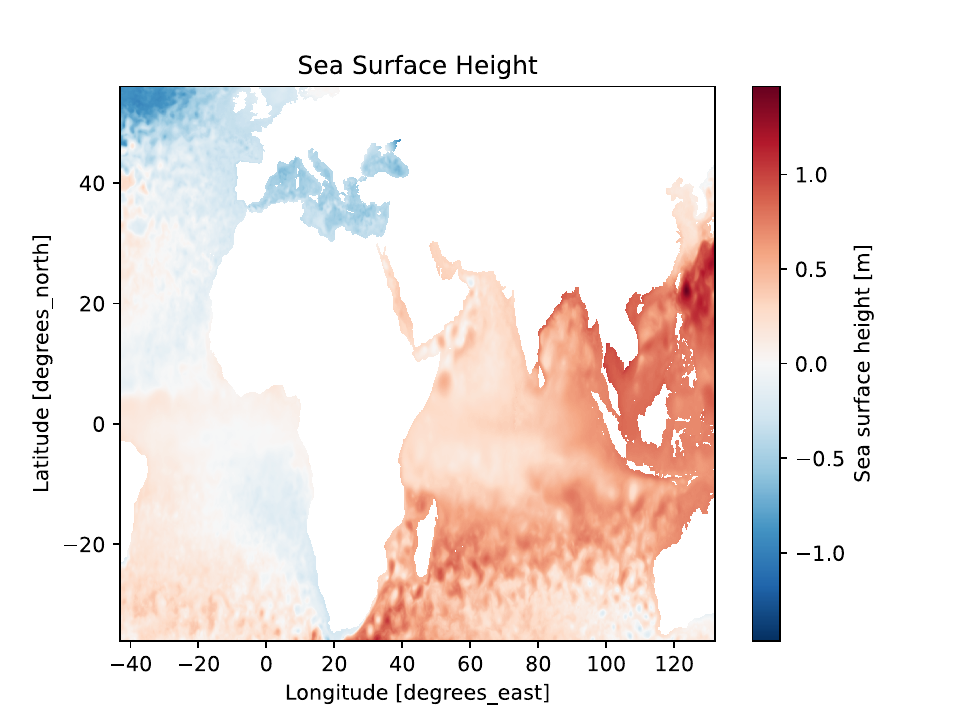}
        \caption{Surface Height}
        \label{fig:surface_height}
    \end{subfigure}
    \begin{subfigure}[b]{0.45\textwidth}
        \centering
        \includegraphics[width=\textwidth]{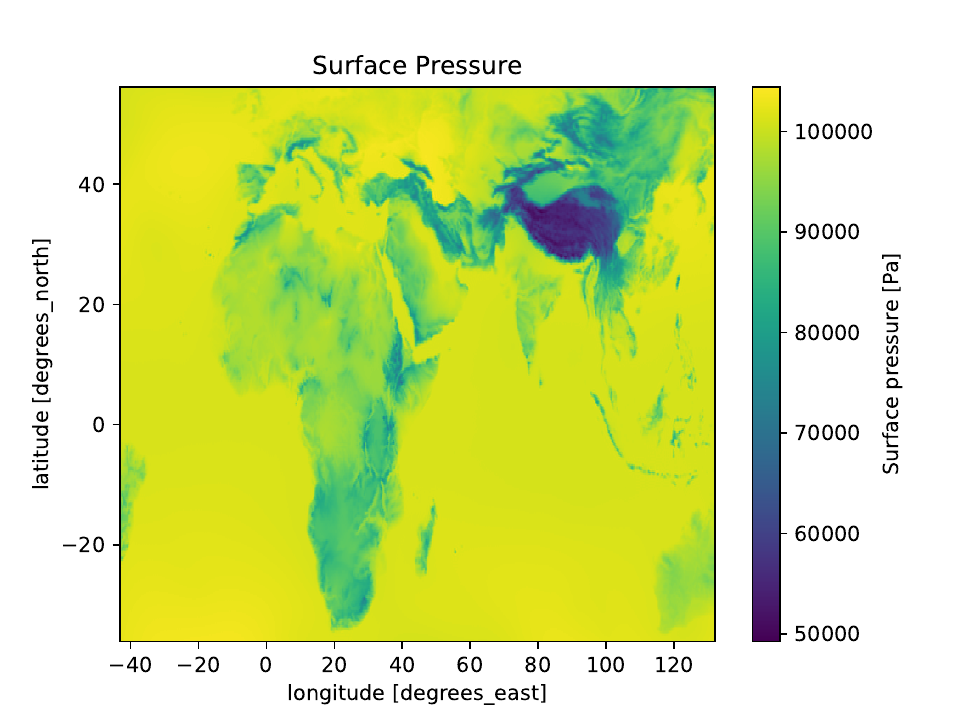}
        \caption{Surface Pressure}
        \label{fig:surface_pressure}
    \end{subfigure}
\end{figure}

\end{document}

%% file: abstract.tex

\begin{frontmatter}

    \begin{abstract}

        There is an increasing push for operational measures to reduce ships' bunker \ac{fc} and carbon emissions, driven by \ac{imo} mandates. Key performance indicators like \ac{eeoi} focus on fuel efficiency. Trim optimization, virtual arrival, and green routing strategies have emerged. The theoretical basis for all these approaches lies in the accurate predictions of \ac{fc} based on its sailing speed, displacement, trim, climate, and sea state. 

        This study utilized 296 voyage reports from a Bulk Carrier Vessel over one year (from November 16, 2021, to November 21, 2022) and 28 parameters, integrating hydrometeorological big data from \ac{cmems} and \ac{ecmwf} with 19 and 61 parameters, respectively, to evaluate if fusing external public data sources may enhance modeling accuracy as well as highlighting most influential parameters affecting \ac{fc}.
        
        The experimental results revealed that there is a large potentioal of \ac{ml} techniques in predicting ship \ac{fc} accurately by using voyage reports and fusing reliable climate and sea data. However, further validation on similar class of vessels remains necessary to confirm the model's generalizability.
 
    \end{abstract}

    \begin{keyword}
        Voyage Optimization, Bulk Carrier, \ac{fc} Modelling, Machine Learning, CMEMS, ERA5
    \end{keyword}

\end{frontmatter}

%% file: acronyms.tex
\section*{Nomenclature}
\renewcommand{\baselinestretch}{0.75}\normalsize
\renewcommand{\aclabelfont}[1]{\textsc{\acsfont{#1}}}
\begin{acronym}[longest]

    \acro{ann}[ANN]{Artificial Neural Network}

    \acro{bbm}[BBM]{Black Box Model}
    \acro{blstm}[BLSTM]{Bidirectional Long Short-Term Memory}

    \acro{cmems}[CMEMS]{Copernicus Marine Environment Monitoring Servic}
    \acro{cii}[CII]{Carbon Intensity Indicator}
    \acro{cfd}[CFD]{Computational Fluid Dynamics}
    
    \acro{dt}[DT]{Decision Tree}
    \acro{ddm}[DDM]{Degree Decimal Minutes}
    \acro{dd}[DD]{Decimal Degrees}
    \acro{dl}[DL]{Deep Learning}
    \acro{dcs}[DCS]{Data Collection System}
    
    \acro{eeoi}[EEOI]{Energy Efficiency Operating Indicator}
    \acro{eu}[EU]{European Union}
    \acro{era5}[ERA5]{Fifth Generation \ac{ecmwf} Atmospheric Reanalysis of the Global Climate}
    \acro{eda}[EDA]{Exploratory Data Analysis}
    \acro{ecmwf}[ECMWF]{European Centre for Medium-Range Weather Forecasts}

    \acro{fc}[FC]{Fuel Consumption}
    \acro{foc}[FOC]{Fuel Oil Consumption}
    
    \acro{ghg}[GHG]{Green House Gas}
    
    \acro{hfo}[HFO]{Heavy Fuel Oil}

    \acro{imo}[IMO]{International Maritime Organization}

    \acro{kpi}[KPI]{Key Performance Indicator}

    \acro{lng}[LNG]{Liquefied Natural Gas}

    \acro{ml}[ML]{Machine Learning}
    \acro{mgo}[MGO]{marine gas oil}
    \acro{marpol}[MARPOL]{International Convention for the Prevention of Pollution from Ships}
    \acro{mrv}[MRV]{Measurement Reporting and Verification}
    \acro{mse}[MSE]{Mean Squared Error}
    \acro{mae}[MAE]{Mean Absolute Error}
    \acro{mlr}[MLR]{Multiple Linear Regression}

    \acro{noaa}[NOAA]{National Oceanic and Atmospheric Administration}

    \acro{ols}[OLS]{Ordinary Least Square}

    \acro{r2}[R2]{Cofficient of Determination}
    \acro{rmse}[RMSE]{Root Mean Squared Error}
    \acro{rr}[RR]{Ridge Regression}
    \acro{rf}[RF]{Random Forest}
    \acro{rpm}[RPM]{Rotation Per Minute}

    \acro{std}[Std]{Standard Deviation}
    \acro{sfoc}[SFOC]{Specific Fuel Oil Consumption}
    \acro{stw}[STW]{Speed Through Water}
    \acro{sog}[SOG]{Speed Over Ground}
    \acro{shap}[SHAP]{SHapley Additive exPlanations}
    \acro{svr}[SVR]{Support Vector Regression}

    \acro{un}[UN]{United Nations}
    \acro{utc}[UTC]{Universal Coordinated Time}
    \acro{us}[US]{United States}

    \acro{vlsfo}[VLSFO]{Very-Low Sulphur Fuel Oil}

    \acro{wbm}[WBM]{White Box Model}

    \acro{xgboost}[XGBoost]{eXtreme Gradient Boosting}

\end{acronym}
\renewcommand{\baselinestretch}{1}\normalsize

%% file: introduction.tex
\section{Introduction}\label{sec:introduction}

Maritime transportation is the driving engine of the global economy, representing 80 \% of the world trade by value~\citep{Unctad2023}. It is the most energy-efficient mode of transportation compared to air, rail, or road, thanks to its carrying capacity and economical fuel type. An average-size cargo shipment is comparable to 2000 trucks, 2500 airplanes, or 225 trains as mentioned by Pena et al.~\citep{penaReviewApplicationsMachine2020}.

Much of the emphasis recently has been as highlighted by Ji et al.,Pavlenko et al.,Stolz et al.~\citep{jiDatadrivenStudyIMO2020, pavlenkoClimateImplicationsUsing, stolzTechnoeconomicAnalysisRenewable2022} on the usage of clean energy such as \ac{vlsfo}, \ac{mgo} and \ac{lng}. 
However, according to the 4th \ac{ghg} study \ac{imo}~\citep{imoFourthIMOGHG2020}, 
most vessels today still uses \ac{hfo}, which is roughly 30\% cheaper compared to other fuel types since \ac{hfo} is the residual product of the oil refinement process for the extracption of gasoline and diesel so-called distillates for use in planes, trucks and cars which leaves it as waste product and this makes it cost-effective~\citep{clearseasMarineFuelsWhat2020}. 

This fuel is used in the ship in four main areas: main engine propulsion, energy saving for main engine, auxiliary propulsion, and energy saving for auxiliary engine as mentioned by Fan et al.~\citep{fanReviewShipFuel}. However, although shipping is a very efficient mode of transport, it has a significant footprint on the climate; this is multiplied as the volume of international trades that rely on shipping is rising. Roughly 3\% of all global \ac{ghg} emissions are caused by maritime transportation~\citep{Unctad2023}. To put this into perspective, that is 1 Billion tons of \ac{ghg} emitted annually to the atmosphere, which is equivalent to the emissions of whole countries such as Mozambique, for example, with 28 million habitats Pena et al.~\citep{penaReviewApplicationsMachine2020}.

To reduce this, \ac{imo} have proposed regulations that pressure the industry to drive change. One of the leavers in the context of \ac{ghg} is the ambition to lower emissions by half by the year 2050 compared to levels of 2008 which was agreed upon in the Paris Agreement on climate change in~\citep{unParisAgreement2015} with the overall target of lowering global temperature to less than 2° globally Pena et al.~\citep{penaReviewApplicationsMachine2020}.

This shared goal puts an urgent need to optimize voyages to meet \ac{imo} goals. All optimization efforts are classified into two categories: design optimization, for example, wind-assisted propellers or hull-fouling, and operational optimization, such as just-in-time arrival, slow-steaming, and green routing.

During operation there exists a strong connection between \ac{eeoi}, \ac{cii} and \ac{fc} Ailong et al.,Zhang et al.,Yapeng et al.~\citep{FAN2021120266, zhang2017multi, HE2021109733} as cited by Fan et al.~\citep{fanReviewShipFuel}. Therefore, the reduction of fuel can result in significant operational cost savings. Hence, optimal \ac{fc} is crucial for realizing this. Also, as pointed out by the latest review paper Ran et al.~\citep{YAN2021102489} as cited by Li et al.~\citep{liDataFusionMachine2022}, the basis of all operational measures for ship bunker fuel savings and emission mitigation is the quantitatively modeling the relationship between \ac{fc} rate and its determinants, including sailing speed, draft/displacement, trim, weather conditions, and sea conditions, but it is not a trivial work.

The accurate predictions of \ac{fc} come down to selecting a suitable model and data that captures all the influences on the bunker fuel. A recent review paper highlighted that Fan et al.~\citep{fanReviewShipFuel} choice of models could be classified into two types: (a) \ac{wbm}, sometimes referred to as physics-based models, (b) \ac{bbm}, sometimes referred to as \ac{ml} models.

White-box models are methods that are based on ship and engine-propeller performance models; see review Ran et al.~\citep{YAN2021102489} and Fan et al.~\citep{fanReviewShipFuel}.
They are based on accurately estimating total ship resistance Zhang et al.~\citep{zhangDeepLearningMethod2024}. The resistance that the ship experiences during sailing could be classified into two types: (a) Calm water resistance and (b) In-service resistance, sometimes referred to as added resistance Fan et al.~\citep{fanReviewShipFuel}.
Calm water resistance refers to the hydrodynamic effects the ship encounters when sailing in undisturbed waters, such as viscous or wave-making resistance, which are generally caused by the ship. In contrast, added resistance is based on the environment. It refers to the hydrodynamics effects on a ship from external factors such as winds, waves, other ships, and obstacles Vinayak et al.,Zhihua et al.~\citep{vinayaketal2021, LIU2020107246} as cited by Zhang et al.~\citep{zhangDeepLearningMethod2024}. The total resistance of a ship is computed by adding both of those components, and the most common methods used are the towing tank and \ac{cfd} Fan et al.~\citep{fanReviewShipFuel}. \ac{wbm}, as the name suggests, are transparent in their structure and interpretable in nature. They rely on physical principles, theoretical frameworks, and explicit mathematical formulations to explain the relation between Ship \ac{fc} and its determinants. They are often used in the ship design phase to manufacture efficient ships.
The parameters used in those classes of models are inter-connected and, therefore, susceptible to the environment, which leads to errors, especially in the case of extreme navigation environments Nan et al.~\citep{WEI2021121036} as cited by Fan et al.~\citep{fanReviewShipFuel}. Additionally, the parameters derived in those formulas are not adjustable, which means they cannot be changed during the voyage nor account for degradation over time in the propulsion systems Fan et al.~\citep{fanReviewShipFuel}.
The performance of the \ac{wbm} is strongly affected by various assumptions Ran et al.~\citep{YAN2021102489} as cited by Fan et al.~\citep{fanReviewShipFuel}.
For example, the components and interactions of the ship resistance are often ignored, resulting in poor applicability of \ac{fc} models Haranen et al.~\citep{haranenWhiteGreyBlackBox} as cited by Fan et al.~\citep{fanReviewShipFuel}

The term "white-box" contrasts with "black-box" models, which do not have transparent internal workings and rely heavily on empirical data and statistical methods without necessarily explaining the underlying physical processes. There is a vast potential for using \ac{ml} methods to model Ship \ac{fc} which can solve the challenges that physics-based models Chen et al.,Lang et al.,Shang et al.~\citep{chenetal2023, lang2023data, shangetal2023} as cited by Zhang et al.~\citep{zhangDeepLearningMethod2024}.
This is because \ac{ml} methods could elucidate the inter-relationship between the measured \ac{fc} and its determinant factors based solely on the observed measurements with the ability to take realistic navigational patterns, ship operational status, weather conditions, and engineering systems specifics Zhang et al.~\citep{zhangDeepLearningMethod2024}.~\ac{ml} excels in decoding intricate patterns, making it particularly advantageous for advanced shipping applications, which is the primary focus of this study.

The remainder of this paper is organized as follows: \cref{sec:literature_analysis} The literature analysis, where there is a narrow focus on the current state of research in terms of Ship \ac{fc} prediction using \ac{ml} approaches, is presented, together highlighting the gaps and contribution of this study. \cref{sec:methodological_approach} cover the methodological approach used, and \cref{sec:results} highlights and discuss the results. Finally, a conclusion of the findings is drawn together with an outlook of future research are presented in \cref{sec:conclusion}.

%% file: literature_analysis.tex
\section{Literature Analysis}\label{sec:literature_analysis}

Our study focuses on modeling \ac{fc} rate (MT/day) of a bulk carrier using \ac{bbm} and fusing operational data with environmental factors. \\
In this regard, the most recent review papers in the space are done by Huang et al.~\citep{huangMachineLearningSustainable2022} and Fan et al.~\citep{fanReviewShipFuel}.  
Huang et al.~\citep{huangMachineLearningSustainable2022} covered applications of \ac{bbm} in different ship phases such as ship design, voyage planning, and operational performance. 
While Fan et al.~\citep{fanReviewShipFuel} analyzed the literature from 2001 to 2021 and focused on applications of \ac{bbm} in operational context only with a comparison of different 
\ac{ml} classifications in various ship operation stages and based on data availability.

To avoid duplication, this section will focus on work that emerged since then that is not covered in the above reviews regarding \ac{fc} prediction by fusing external data sources. 

Xie et al.~\citep{xieFuelConsumptionPrediction2023} compared between \ac{bbm} and \ac{wbm} and proved \ac{bbm} superiority. They developed two ensemble models, namely \ac{rf} and 
\ac{xgboost} models for an oil tanker vessel based on eight features. They fused external public weather data from \ac{ecmwf} considering only two parameters: wave height and wave direction, and then applied a novel data cleaning method, namely Kwon formulas, to remove noise from the data. The results revealed that Kwon formulas are a promising approach to improving data quality. However, it could lead to idealized data used for training, which would lead to poor real-world applicability. In addition, 
they used model test data, a scaled-down representation that is good for predictions in the ship design phase but not in the operational performance context.

In contrast, Zhou et al.~\citep{zhouPredictingShipFuel2023} used in-situ data, which is full-scale data based on the natural environment, e.g., open sea. 
Their model is a 2-stage sequential model based on a fishing vessel. In the first stage, a prediction of \ac{stw} is made based on \ac{sog} and environmental conditions and then using 
\ac{stw} predictions as input for predicting \ac{foc} in the second stage. The final selected model is based on \ac{rf} architecture. Their prediction output is \ac{foc}, which is different than \ac{fc}.
In the sense that \ac{foc} exclusively applies to oil-based fuel types such as \ac{hfo}, which don't take into consideration greener fuel types. This reduces its readiness for future-proof transition to other types of fuels. 
Also, they fused environmental data from \ac{noaa} data store, which lacks global coverage since it's more suited for voyages within \ac{us} coastal waters. In addition, it lacks standardized APIs that streamline data acquisition, making it impractical for real-world applications. 

Agand et al.~\citep{agandFuelConsumptionPrediction2023} investigated different models such as \ac{mlr}, \ac{ann} and ensemble models. They found that ensemble models performed the best, which aligns with Xie et al.~\citep{xieFuelConsumptionPrediction2023} findings. 
Their model is based on a passenger ferry within a 2-year span and in-service data, which is high-frequency, high-precision data that reflects near real-time changes. 
The model predicts hourly \ac{fc}, which makes it, to some extent, suitable for real-time applications such as route optimization. However, they didn't consider localized environmental factors.

Haoqing et al.~\citep{WANG2023104361} attempted to improve \ac{dl} interpretability by building \ac{ann} that integrates physics and domain knowledge in order to have an explainable model that balances 
the trade-off between interpretability and accuracy. They achieved that by using \ac{shap}, a method developed by Scott et al.~\citep{lundberg2017unified} based on game theory. 
The model was trained using publically accessible data over two months from a ferry vessel originating from a study published in Petersen et al.~\citep{petersen2012machine} as cited by Haoqing et al.~\citep{WANG2023104361}. 
The model was trained on very old data, roughly 12 years or more old. This poses a risk of applicability moving forward since the data is no longer relevant, as it won't reflect current shipping practices or take into consideration technologies of new ship builds. 

\subsection{Gaps and Contribution}\label{sec:gaps_and_contribution}

To the best of our knowledge, the most recent and closest work to our study is the work done by Zhang et al.~\citep{zhangDeepLearningMethod2024} and Du et al.~\citep{duDataFusionMachine2022a}. 
Zhang et al.~\citep{zhangDeepLearningMethod2024} developed a model for bulk carriers. Their model is based on \ac{dl} technique, namely \ac{blstm}, and compared it to traditional \ac{ml} methods. 
They found that \ac{blstm} provides better accuracy compared to traditional \ac{dl} and \ac{ml}. However, the difference was not significant enough to justify the 
complexity associated with the usage of such a model in practice. Additionally, they are different to our study that they require high resolution input data and the predictions 
are made per voyage instead of daily predictions. 

Another study closer to ours that was to some extent used as a reference to our study is Du et al.~\citep{duDataFusionMachine2022a}. 
The study's main objective was to fuse environmental conditions to AIS data, and the final model selected is \ac{xgboost}. They are different from our study; however, since they applied it to containerships, our work attempts to apply it in a different context, namely on bulk carriers, which is different due to the data structure and nature of sailing.

Looking at the literature landscape concerning modeling \ac{fc} using \ac{bbm} while fusing external public and reliable meteorological factors, 
To the best of our knowledge, there are not many recent studies that have studied this extensively. Some studies like the work done by Xie et al.~\citep{xieFuelConsumptionPrediction2023} attempted to fuse public environmental parameters but only considering two parameters namely wave heights and direction and therefore no 360° view of the ship envionment was considered.  
This finding is also supported by latest findings by Li et al.~\citep{liDataFusionMachine2022}. 

This raises the overarching question: "Does fusing meteorological factors in ship \ac{fc} modeling gives a better and more accurate model?"

\clearpage

%% file: methodological_approach.tex
\section{Methodological Approach}\label{sec:methodological_approach}
\begin{figure}[!ht]
    \centering
    \includegraphics[width=1\linewidth]{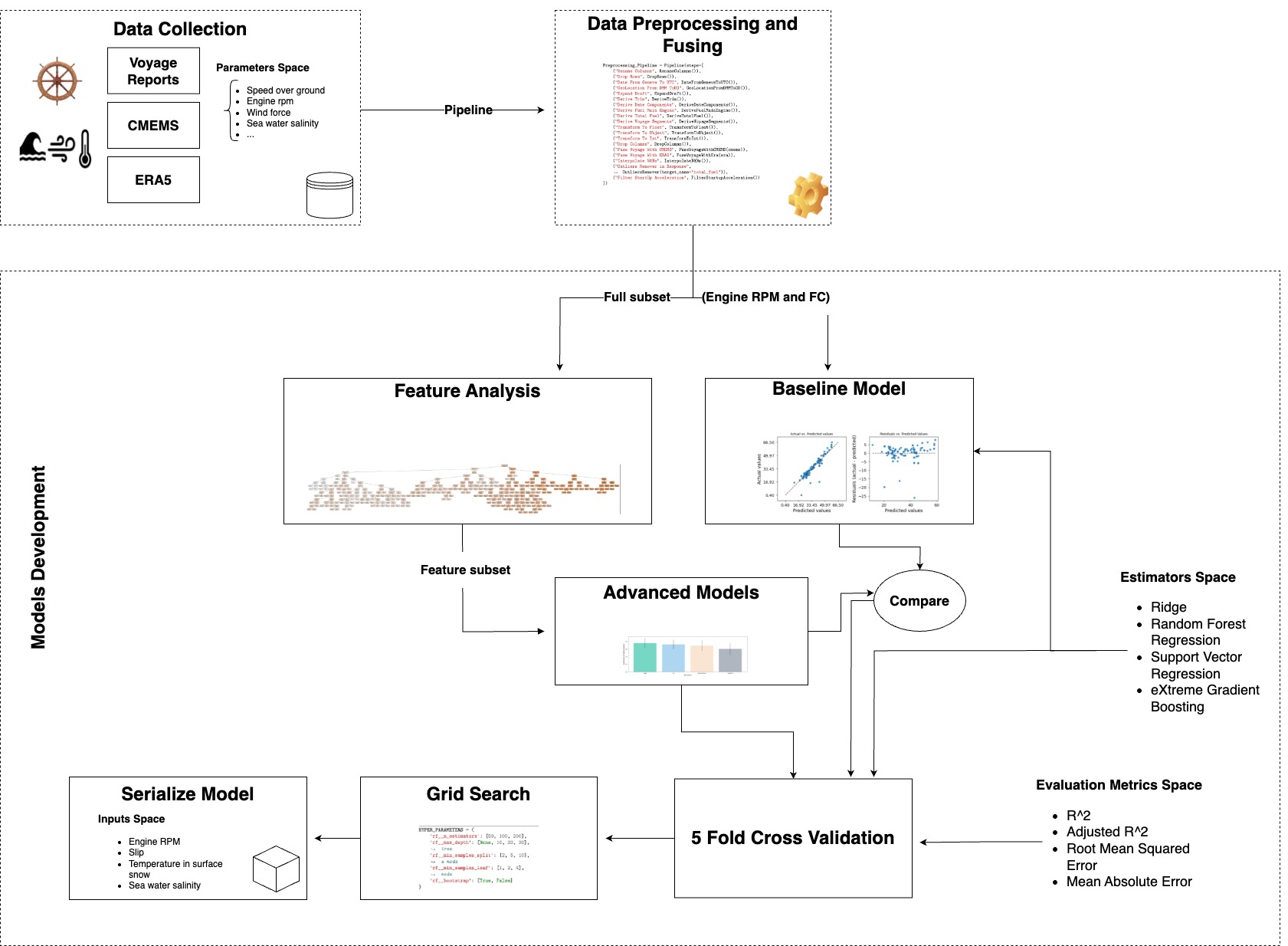}
    \caption{Illustration of Methodological Approach}
    \label{fig:illustration_of_methodological_approach}
\end{figure}

The methodological approach in this study is described in the subsection below and visualized in \cref{fig:illustration_of_methodological_approach}. 

\subsection{Data Collection}\label{sec:data_collection}
The datasets used to conduct this study are all based on three public sources: a) Voyage Reports b) \ac{cmems} and c) \ac{era5}. 

\subsubsection{Voyage Reports}\label{sec:voyage_reports}
Voyage reports data, sometimes also referred to as noon reports, are data that are captured manually and reported by the chief engineer daily at noon, hence the name, and then sent by the ship master to the shipping company and shore management to report operational and performance status of the vessel as mentioned by Ran et al.~\citep{YAN2021102489}. These reports are mandated by \ac{imo} to meet regulatory requirements such as \ac{dcs}, which mandate since 2019, all ships over 5,000 gross tons to report their daily aggregated fuel consumption as mentioned by \ac{imo}~\citep{imoIMODataCollection2023}. 

Noon reports contain navigational information (e.g., latitude and longitude), operational information (e.g., speed over ground, engine rpm), and weather information (e.g., wind speed and swell force). A raw sample of the reports used in the study is shown in \cref{tab:voyage_report_raw_samples_subset} and for a full list of the parameters retrieved together with their unit of measurements see \ref{tab:voyage_reports_raw_parameters_and_unit_of_measure} in appendix. 

\clearpage
\input{voyage_report_raw_samples_subset}

\begin{figure}
    \centering
    \includegraphics[width=1\linewidth]{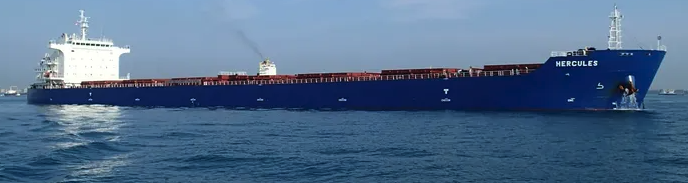}
    \caption{Hercules Vessel}
    \label{fig:hercules}
\end{figure}

The reports originated from a bulk carrier vessel named Hercules, shown in \cref{fig:hercules}. 
For an overview of the vessel particulars and power specifications, see \cref{tab:hercules_particulars,tab:hercules_power_specifications}. 

\input{hercules_particulars}

The timespan of data used is from 16 November 2021 to 21 November 2022 in daily frequency, which amounts to 296 samples and 28 parameters. 

Since voyage reports are logged manually, specific measurements generally will have human errors, specifically for weather and sea parts of the measurements recorded since they are gauged. Moreover, the weather parameters logged in these reports are limited to a handful of parameters and also missing many signals. This motivated the inspection of other more reliable and quality weather and sea state data that could be fused to produce better models. 

Our investigation revealed that the highest quality data vendors that fulfill this are using \ac{cmems} as well as the \ac{era5}, which is, according to Li et al.~\citep{liDataFusionMachine2022}, what windy~\citep{windy} uses internally, which is an application that is widely adopted for assessing maritime navigation in real-time by providing reliable weather forecasts in the open ocean.

\input{hercules_power_specifications}

\subsubsection{\ac{cmems}}\label{sec:cmems}

\ac{cmems} provides high-quality data on the ocean, which includes information on the state of the Blue (physical), White (sea ice), and Green (biogeochemical) ocean on a global and regional scale. It integrates satellite observations, in-situ measurements, and advanced numerical models. It offers information specifically on the ocean, such as currents, sea level, salinity, and biogeochemical parameters ~\citep{copernicusmarineservice}.

It has a whole host of different products, and the product selected for this study is Global Ocean Physics Analysis and Forecast~\citep{cmems2024global}, which was inspired by the study Li et al.~\citep{liDataFusionMachine2022}.

The product comprises five different datasets, each with different parameters. \cref{tab:global_ocean_physics_analysis_and_forecast_data_description,tab:global_ocean_physics_analysis_and_forecast_raw_samples} describe the metadata and a sample of the raw data used. For details on the parameters together with their unit of measurements see \cref{tab:cmems_raw_parameters_and_unit_of_measure} in appendix. 

Since \ac{cmems} focuses on oceanic data, another data source was needed to be fused, which is \ac{era5}, which completes environmental data by covering atmospheric reanalysis that includes information on the climate such as wind speed, wind direction, and temperatures. 

\input{global_ocean_physics_analysis_and_forecast_data_description}

\input{global_ocean_physics_analysis_and_forecast_raw_samples}

\subsubsection{\ac{era5}}\label{sec:era5}

\ac{era5} is a global atmospheric reanalysis dataset produced by \ac{ecmwf}. It provides hourly data on the climate from 1959 to the present, combining model data with observations Hersbach et al.~\citep{hersbach2023era5}.

The data obtained were selected from the product, hourly data on single levels from 1940 to the present, which was inspired by the study ~\citep{liDataFusionMachine2022}. This provided 61 parameters related to the climate that were used to test if they contributed to improving the overall performance.

A short description of the dataset used together with a sample on the raw measurements are depicted in~\cref{tab:era5_hourly_on_single_levels_data_description,tab:era5_hourly_on_single_levels_raw_samples}.
For details on the used parameters their unit of measurements see \cref{tab:era5_raw_parameters_and_unit_of_measure} in appendix.

\clearpage

\input{era5_hourly_on_single_levels_data_description}

\input{era5_hourly_on_single_levels_raw_samples}

Data preprocessing significantly contributes to the generalization performance of a supervised \ac{ml} algorithm~\citep{xieFuelConsumptionPrediction2023}, and therefore, this was carried out next.

\subsection{Data Preprocessing and Fusing}\label{sec:data_preprocessing_and_fusing}

To achieve the best results, it is also crucial to carefully balance the need to clean the data with maintaining the integrity of the raw data. Hence, having a consistent pipeline was crucial to avoid data leakage during training and ensure reproducibility. Therefore, Sklearn pipelines have been used, which are sequential lists of steps that could be applied to the data. Each step could be a transformation or an estimator. This standardized the process of building models, which ensures consistency and reproducibility.

\begin{figure}
    \centering
    \includegraphics[width=1\linewidth]{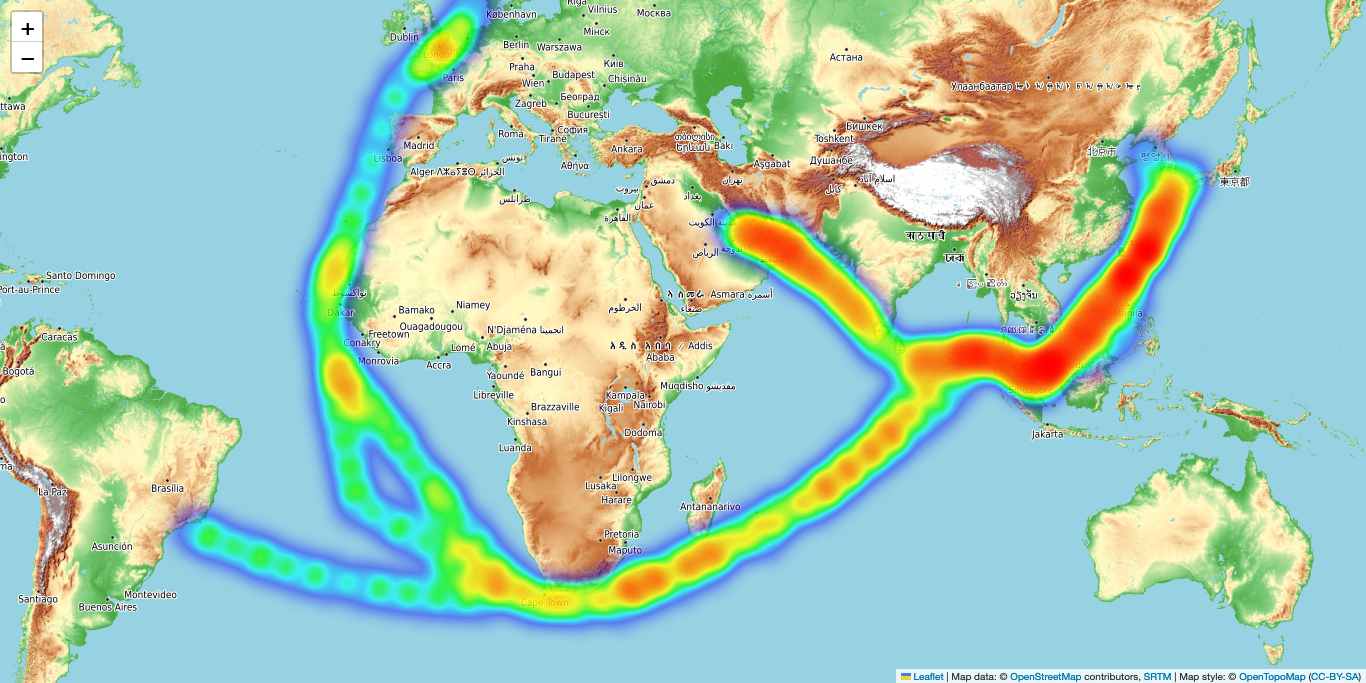}
    \caption{Waypoints Visited}
    \label{fig:waypoints_visited_frequency}
\end{figure}

At first, the data were converted from Excel to pandas data frames so that the tools could be applied to them. Then, some columns were renamed to have more meaningful names. Then, any observation involving a shift in state in any of the following (Loading Operation, Anchor, Bunkering Operation, Discharging Operation, VSL Anchored, and Drifting) was removed from the data since it had empty values across all measurements. Next, some alignments were done to map the noon reports to the environmental sources. These alignments included that the dates have been converted from Geneva to \ac{utc}. The geographical positions have been converted from \ac{ddm} to \ac{dd}.
Then, some measurements have been derived from the raw measurements. This includes expanding the draft to a forward draft and an aft draft. Then, the trim from the draft values is computed. Then, the date parameters were expanded into year, month, day, and season based on whether the ship's location was in the north or south hemisphere. Then, the main engine's fuel was derived by adding two different fuel types, mainly ultra-low sulfur fuel and gas fuel. Ultimately, this rolled up to total fuel, which is based on adding fuel used in the main engine, the boilers, and any auxiliary systems. Lastly, the voyage segments were derived, which is the clustering of different port-to-port trips made by the vessel.
Some types of conversions were necessary to accommodate machine learning. This involved transforming some columns, such as total fuel or propeller slip, to floats while converting other measurements to integers or objects.
Next, \ac{cmems} and \ac{era5} data were fused to the voyage data. Since \ac{era5} data are in high resolution hourly, it was averaged to daily noons, reducing the overall size of the data from 0.5 TB to 10GB. The fusing is then done by extracting climate and ocean parameters and matching them with the voyage time and geographical location.
Lastly, linear interpolation with backward fill was carried out to fill for missing observations where climate and ocean measurements along ship voyages were missing. Linear interpolation fills missing values based on the mean of the upper and lower values. Then, the removal of observations and the dropping of some measurements were necessary. This involved the removal of columns that add no value to the predictive power, such as constant parameters or parameters with lots of missing values greater than 5\%, which only added noise. Examples include current speed over the ground since it is correlated with an average speed over the ground or next port, etc. The outliers in the response variable were identified in roughly 3-5 observations and eliminated, and this proved to improve the overall predictive power, as discussed in \cref{sec:results}. Finally, based on the observation made on fuel consumption relation with speed, which is discussed in \cref{sec:results}, the start-up acceleration instances where the speed is high but fuel consumption is lower than usual for bulk carrier of this size have been identified which aligns with Xie et al.~\citep{xieFuelConsumptionPrediction2023} and a threshold was set to 15 ton/day as a minimum where selected based on a typical bulk carrier of this particulars \cref{tab:hercules_particulars}.
The final data used after preprocessing for training were 266-noon reports and 94 parameters, including the response variable. The historical visited waypoints by the vessel is shwon in \cref{fig:waypoints_visited_frequency}. 
It illustrates the spatial distribution of the case study vessel's operational frequency across global maritime routes. Areas of higher intensity (red and orange regions) indicate zones with more frequent vessel presence, while cooler colors (green to blue) represent lower visitation frequencies. The heatmap clearly reveals the vessel’s primary operational corridors, which align with major global shipping lanes, including:  

\begin{itemize}
    \item The Strait of Malacca and South China Sea, indicating intensive activity in Southeast Asian waters
    \item The Bay of Bengal and Arabian Sea, with a prominent route extending through the Suez Canal into the Mediterranean Sea
    \item Southern passages around Africa (Cape of Good Hope) and connectivity to the South Atlantic
    \item Eastward trajectories suggest routes to Australia and the Western Pacific.

\end{itemize}

This spatial footprint highlights the vessels integration in key international trade routes and underscores the relevance of the selected case study vessel for analyzing fuel consumption and voyage dynamics in real-world global maritime operations.
A sample code from the source code used for the data transformation Source Code \ref{src:preprocessing_pipeline} demonstrate the applied transformational steps in the pipeline.

\begin{figure}
\begin{lstlisting}[
        language=Python,
        firstline=56,
        lastline=76,
        caption={Preprocessing Pipeline},
        label={src:preprocessing_pipeline}
    ]
Preprocessing_Pipeline = Pipeline(steps=[
    ("Rename Columns", RenameColumns()), 
    ("Drop Rows", DropRows()), 
    ("Date From Geneva To UTC", DateFromGenevaToUTC()), 
    ("GeoLocation From DMM ToDD", GeoLocationFromDMMToDD()),
    ("Expand Draft", ExpandDraft()), 
    ("Derive Trim", DeriveTrim()), 
    ("Derive Date Components", DeriveDateComponents()), 
    ("Derive Fuel Main Engine", DeriveFuelMainEngine()), 
    ("Derive Total Fuel", DeriveTotalFuel()), 
    ("Derive Voyage Segments", DeriveVoyageSegments()), 
    ("Transform To Float", TransformToFloat()),
    ("Transform To Object", TransformToObject()),
    ("Transform To Int", TransformToInt()),
    ("Drop Columns", DropColumns()), 
    ("Fuse Voyage With CMEMS", FuseVoyageWithCMEMS(cmems)), 
    ("Fuse Voyage With ERA5", FuseVoyageWithEra(era)), 
    ("Interpolate NANs", InterpolateNANs()), 
    ("Outliers Remover in Response", OutliersRemover(target_name="total_fuel")), 
    ("Filter StartUp Acceleration", FilterStartupAcceleration())
])
\end{lstlisting}
\end{figure}

\clearpage

\subsection{Models Development}\label{sec:models_development}

The model development procedure were divided into three parts: a) Baseline Model, b) Feature Analysis and c) Advanced Model. 

\subsection{Baseline Model}\label{sec:baseline_model}

The baseline model assesses overall performance using only one main predictor which is: Engine \ac{rpm}, one of the most essential variables in predicting ship \ac{fc}. 
This is more relevant than using \ac{sog} as discussed in \cref{sec:results} and also in alinment with the findings by ~\citep{Uyanik2022}.

This predictor was used to produce a baseline model without taking into consideration any environmental parameters nor operational factors in order to assess the extend 
to which variations in the \ac{fc} could be explained. For that multiple estimators with default settings from various categories such as: Linear, Tree Based, Support Vector Regression and Ensembles have been deployed. 
The estimators considered are shown in Source Code \ref{src:ml_estimators}.  

\clearpage

\begin{figure}
\begin{lstlisting}[
        language=Python,
        firstline=17,
        lastline=30,
        caption={ML Estimators},
        label={src:ml_estimators}
    ]
PIPELINES = {
    'ridge': Pipeline([
        ('estimator', Ridge())
    ]),
    'randomforest': Pipeline([
        ('estimator', RandomForestRegressor())
    ]),
    'svr': Pipeline([
    ('estimator', SVR(kernel="linear"))
    ]),
    'xgboost': Pipeline([
        ('estimator', XGBRegressor())
    ])
}
\end{lstlisting}
\end{figure}

\subsubsection{\ac{rr}}\label{sec:rr}

\ac{rr} is a type of linear regression models also referred to as Tikhonov Regularization and developed by Hoerl et al.~\citep{hoerl1970ridge}
which is an addition of the widely known \ac{ols} regression where it introduces penalization propotional to the square of the coefficient to address any 
multicollinearity and overfitting cases. The mathemtical formulation for the cost function is depicted in \cref{eq:ridge_cost}.  

\begin{equation}
    \min_{\beta} \| y - X\beta \|_2^2 + \lambda \|\beta\|_2^2
\label{eq:ridge_cost}
\end{equation}
    
Where:
\begin{itemize}
    \item \( \| y - X\beta \|_2^2 = \sum_{i=1}^n (y_i - X_i \beta)^2 \): Residual sum of squares.
    \item \( \|\beta\|_2^2 = \sum_{j=1}^p \beta_j^2 \): L2 regularization term.
    \item \( \lambda \): Regularization parameter controlling the trade-off between minimizing the residual sum of squares and shrinking the coefficients.
\end{itemize}

\subsubsection{\ac{rf}}\label{sec:rf}

\ac{rf} is an ensemble learning method. It is an enhancement to \ac{dt} which originated by Breiman et al.~\citep{breiman2001random}. 
It is an ensemble in the sense that it combines multiple \ac{dt} to improve accuracy and reduce overfitting. 
It operates by constructing multiple trees and then outputing the mean prediction for regression tasks. 
The mathemtical formulation of the cost function is depicted in \cref{eq:rf_cost}. 

\begin{equation}
    \mathcal{L} = \frac{1}{N} \sum_{i=1}^{N} \left( y_i - \frac{1}{T} \sum_{t=1}^{T} \hat{y}_{i,t} \right)^2
    \label{eq:rf_cost}
    \end{equation}
    
Where:
\begin{itemize}
    \item \( N \): Number of samples.
    \item \( T \): Number of trees in the forest.
    \item \( y_i \): Actual value of the \( i \)-th sample.
    \item \( \hat{y}_{i,t} \): Prediction for the \( i \)-th sample by the \( t \)-th tree.
    \item \( \frac{1}{T} \sum_{t=1}^{T} \hat{y}_{i,t} \): Average prediction across all trees.
\end{itemize}

\subsubsection{\ac{svr}}\label{sec:svr}
\ac{svr} originally developed by Harris et al.~\citep{drucker1997svr}. The working principle is the transformation of data into higher-dimentional space using a kernel 
and then finding a hyperplane that best maximize marginn between different classes then the margin of tolerance is searched. 
The mathemtical formulation for the cost function is depicted in \cref{eq:svr_cost}.

\begin{equation}
\min_{\mathbf{w}, b, \xi^+, \xi^-} \frac{1}{2} \|\mathbf{w}\|^2 + C \sum_{i=1}^n (\xi_i^+ + \xi_i^-)
\label{eq:svr_cost}
\end{equation}

Subject to the following constraints:
\begin{equation}
\begin{aligned}
y_i - (\mathbf{w} \cdot \mathbf{x}_i + b) &\leq \epsilon + \xi_i^+, \quad i = 1, \dots, n, \\
(\mathbf{w} \cdot \mathbf{x}_i + b) - y_i &\leq \epsilon + \xi_i^-, \quad i = 1, \dots, n, \\
\xi_i^+, \, \xi_i^- &\geq 0, \quad i = 1, \dots, n.
\end{aligned}
\label{eq:svr_constraints}
\end{equation}

\begin{itemize}
    \item \( \mathbf{w} \): The weight vector defining the regression hyperplane.
    \item \( b \): The bias term in the hyperplane equation \( f(\mathbf{x}) = \mathbf{w} \cdot \mathbf{x} + b \).
    \item \( \epsilon \): The \emph{insensitivity margin} parameter, which defines a tolerance zone (or \(\epsilon\)-tube) around the hyperplane where deviations are not penalized.
    \item \( \xi_i^+ \): Slack variable representing the amount by which a predicted value exceeds the upper bound of the \(\epsilon\)-tube.
    \item \( \xi_i^- \): Slack variable representing the amount by which a predicted value falls below the lower bound of the \(\epsilon\)-tube.
    \item \( C \): Regularization parameter that controls the trade-off between the model's complexity (\( \frac{1}{2} \|\mathbf{w}\|^2 \)) and the penalty for deviations outside the \(\epsilon\)-tube (\( \sum_{i=1}^n (\xi_i^+ + \xi_i^-) \)).
    \item \( y_i \): The true target value for the \(i\)-th data point.
    \item \( \mathbf{x}_i \): The feature vector for the \(i\)-th data point.
    \item \( n \): The total number of data points in the training set.
\end{itemize}

\subsubsection{\ac{xgboost}}\label{sec:xgboost}

\ac{xgboost} is a type of both ensemble and gradient boosting developed by Tianqi et al.~\citep{chen2016xgboost}. Its main working principle relies on building multiple \ac{dt} similar to 
\ac{rf} but it differs that its based on boosting where trees are trained sequentially to correct the errors of its predecessor unlike \ac{rf} which is based on bagging i.e bootstap aggregation. 
The mathemtical formulation for its cost function is depicted in \cref{eq:xgboost_cost}.

\begin{equation}
\mathcal{L}(t) = \sum_{i=1}^n \ell(y_i, \hat{y}_i^{(t)}) + \sum_{k=1}^t \Omega(f_k),
\label{eq:xgboost_cost}
\end{equation}

where:
\begin{itemize}
    \item \( \ell(y_i, \hat{y}_i^{(t)}) \): Loss function for the \(i\)-th data point (e.g., Mean Squared Error or Mean Absolute Error).
    \item \( \Omega(f_k) = \gamma T + \frac{1}{2} \lambda \|\mathbf{w}\|^2 \): Regularization term for the \(k\)-th tree, where:
    \begin{itemize}
        \item \( T \): Number of leaves in the tree.
        \item \( \gamma \): Regularization parameter for leaf nodes.
        \item \( \lambda \): L2 regularization parameter for leaf weights.
        \item \( \mathbf{w} \): Vector of leaf weights.
    \end{itemize}
    \item \( t \): Current boosting iteration.
    \item \( n \): Number of data points.
\end{itemize}

\subsubsection{Validation of the methods}\label{sec:validation_of_the_methods}

The estimators have been evaluated using cross-validation technique more specifically using 5-fold cross validation. 
The basic principle behind this technique is that data split randomly into k subsets/folds and then the model is trainined in k-1 while getting tested in remaining folds till k times is reached.
This make sure that no overfitting or underfitting phenomenas occur when reporting the final error metrics as its also reported as an average across all folds. 

The technique became prominant in \ac{ml} space for validating models by Stone~\citep{stone1974cross} and its mathemtical formulation is shown below:

Let the dataset \( D \) consist of \( n \) data points, and we split it into \( k \) subsets or folds: 
\begin{equation}
    D = \{D_1, D_2, \dots, D_k\}
\end{equation}
where \( D_i \) is the \( i \)-th fold. 

For each fold \( i \), we train the model on the data from all folds except the \( i \)-th fold, and test it on the \( i \)-th fold. That is, the training data is 
\begin{equation}
    D_{\text{train}} = D \setminus D_i, \quad \text{and the test data is } D_{\text{test}} = D_i.
\end{equation}

The error for fold \( i \) is given by:
\begin{equation}
    \mathcal{E}_i = \frac{1}{|D_i|} \sum_{x_j \in D_i} \mathcal{L}(f(x_j), y_j)
\end{equation}
where:
\begin{itemize}
    \item \( \mathcal{L} \) is the loss function (e.g., Mean Squared Error for regression)
    \item \( f(x_j) \) is the predicted value for the data point \( x_j \)
    \item \( y_j \) is the true value for \( x_j \).
\end{itemize}

The overall cross-validation error is the average error across all \( k \) folds:
\begin{equation}
    \mathcal{E}_{\text{CV}} = \frac{1}{k} \sum_{i=1}^{k} \mathcal{E}_i.
\end{equation}
\clearpage
The models are evaluated using the following performance and error metrics: 
\begin{itemize}
    \item \ac{r2}
    \item Adjusted \ac{r2}
    \item \ac{rmse} 
    \item \ac{mae}
\end{itemize}

\subsubsection{R2 and Adjusted R2}\label{sec:r2_and_adjusted_r2}
\ac{r2} also known as coefficient of determination developed by Francis~\citep{galton1886regression} and it represents the variance in the dependent variable 
that is explained by the independent variables. The adjusted \ac{r2} is a modified version of \ac{r2} developed by George~\citep{mcnemar1978statistical}. Its similar to 
\ac{r2} with slight difference that it take into account the number of predictors in the model so that 
adding more independent variables dont purely inflate \ac{r2}. It works by penalizing the model for adding more independent variables 
that dont contribute with additional information. 

The mathematical formulation for both performance metrics are shown in \cref{eq:r2,eq:r2_adj}.

\begin{equation}
    R^2 = 1 - \frac{\sum_{i=1}^{n} (y_i - \hat{y}_i)^2}{\sum_{i=1}^{n} (y_i - \bar{y})^2}
\label{eq:r2}
\end{equation}

where:
\begin{itemize}
    \item \(y_i\) is the actual value of the \(i\)-th observation.
    \item \(\hat{y}_i\) is the predicted value for the \(i\)-th observation
    \item \(\bar{y}\) is the mean of the actual values.
    \item \(n\) is the number of data points.
\end{itemize}
\clearpage
The formula for Adjusted \(R^2\) is given by:

\begin{equation}
    \bar{R}^2 = 1 - \left(1 - R^2\right) \frac{n-1}{n-p-1}
\label{eq:r2_adj}
\end{equation}

where:
\begin{itemize}
    \item \(n\) is the number of observations.
    \item \(p\) is the number of predictors in the model.
\end{itemize}

\subsubsection{RMSE and MAE}\label{sec:rmse_and_mae}
\ac{rmse} measures the average magnitude of the errors between the predicted and observed values. 
Its first popularized by Willmott et al.~\citep{willmott1985rmse} and it gives an idea of how concentrated the 
data is around the line of best fit. The mathematical formulation for both error metrics are shown in \cref{eq:rmse,eq:mae}.

\begin{equation}
    \text{RMSE} = \sqrt{\frac{1}{n} \sum_{i=1}^{n} (y_i - \hat{y}_i)^2}
    \label{eq:rmse}
\end{equation}

where:
\begin{itemize}
    \item \(y_i\) is the actual value of the \(i\)-th observation,
    \item \(\hat{y}_i\) is the predicted value for the \(i\)-th observation,
    \item \(n\) is the total number of data points.
\end{itemize}

\ac{mae} is a widely used error metric that concentrate on measuring the average of absolute
errors. Huber et al.~\citep{huber1964robust} was the first to put the groundwork for error metrics based on absolute deviations. 
Its less sensitive to extreme values as compared to \ac{rmse} where it average squared errors by squaring the differences and it gives more weight to larger errors. 

\begin{equation}
    \text{MAE} = \frac{1}{n} \sum_{i=1}^{n} |y_i - \hat{y}_i|, \quad 
    \label{eq:mae}
\end{equation}

where:
\begin{itemize}
    \item $n$ is the total number of data points,
    \item $y_i$ is the actual value of the $i$-th data point,
    \item $\hat{y}_i$ is the predicted value for the $i$-th data point,
    \item $|y_i - \hat{y}_i|$ is the absolute error for the $i$-th data point.
\end{itemize}

\subsection{Features Analysis}\label{sec:feature_analysis}

Next, more features needed to be exaimined in order to evaluate its relevance for predicting \ac{fc}. \ac{rf} have been used to evaluate and quantify feature importance. 
Random Forest calculates feature importance by evaluating how much each feature contributes to reducing the impurity 
using Gini impurity in the decision trees. For each feature, measure the total decrease in Gini impurity that results from splitting on this feature, 
averaged over all trees. For mathematical formulation see \cref{eq:feature_importance}. 

\begin{equation}
    \text{Importance of feature } j = \sum_{t=1}^{T} \frac{N_t}{N} \cdot \left( \text{Impurity before split}_t - \text{Impurity after split}_t \right)
    \label{eq:feature_importance}
\end{equation}

\begin{itemize}
    \item $T$ is the total number of trees in the Random Forest.
    \item $N_t$ is the number of samples in tree $t$.
    \item $N$ is the total number of samples in the dataset.
    \item $\text{Impurity before split}_t$ is the impurity of node before splitting (e.g., Gini impurity for classification).
    \item $\text{Impurity after split}_t$ is the impurity of node after splitting (again, Gini impurity or other impurity measure).
    \item The product $\frac{N_t}{N}$ weights the impurity change by the proportion of samples in each tree.
\end{itemize}

Statistical analysis reveals that the existing MLMs may be sensitive to the varying numbers of influencing factors (inputs), apparently ranging from 7 to 75 Chen et al.~\citep{chenetal2023}
A handful of the inputs variables have been identified to contribute to model accuracy as discussed in \cref{sec:results}. Next those identified features have been used to build advanced models that are then compared against baseline. 

\subsection{Advanced Model}\label{sec:advanced_model}
An advanced multivariate model was developed and compared against a baseline model in which all the selected predictors were utilized. This has revealed a tangible improvement in overall performance which is discussed in \cref{sec:results}

Then grid search was deployed to search best parameters to be used for the final model. 
Grid search is a technique that is evolved from optimization domain and made popularized by Pedregosa et al.~\citep{pedregosa2011scikit}. The general idea is that it tries to search through manually specified subsets of hyperparameter space 
to find an optimal combination that gives the best model performance. For the mathematical formulation see \cref{eq:grid_search}

\begin{equation}
    \Theta^* = \arg\max_{\Theta \in G} J\big(f(\Theta), D_{\text{val}}\big)
    \label{eq:grid_search}
    \end{equation}
    
    where:
    
    \begin{itemize}
        \item \( \Theta \): Set of hyperparameters (e.g., learning rate, regularization term).
        \item \( G \): Grid of possible hyperparameter combinations.
        \item \( J \): Objective function, such as validation accuracy or error.
        \item \( f(\Theta) \): Model parameterized by the hyperparameters \( \Theta \).
        \item \( D_{\text{val}} \): Validation dataset used to evaluate the model's performance.
        \item \( \Theta^* \): Optimal hyperparameters that maximize or minimize \( J \).
    \end{itemize}

\clearpage

%% file: voyage_report_raw_samples_subset.tex
\begin{table}[h!]
       \centering
       \caption{Voyage Report Raw Samples Subset}
       \resizebox{\textwidth}{!}{%
       \begin{tabular}{lll}
       \toprule
       {} &             \textbf{position} & \textbf{current\_speed\_over\_ground} \\
       \textbf{date}                &                      &                           \\
       \midrule
       16 NOV 2021 1200LT  &  02-16.0N\textbackslash n101-52.5E &                      13.6 \\
       17 NOV 2021 1200LT  &  05-57.0N\textbackslash n097-39.5E &                      13.5 \\
       18 NOV 2021 1200LT  &  06-15.4N\textbackslash n092-16.2E &                      13.7 \\
       19 Nov 2021\textbackslash n1200LT &  05-47.4N\textbackslash n086-37.4E &                      13.5 \\
       \bottomrule
       \end{tabular}%
       }
       \label{tab:voyage_report_raw_samples_subset}
\end{table}

%% file: hercules_particulars.tex
\begin{table}[h!]
    \centering
    \caption[Hercules Particulars]{Hercules Particulars. Source ~\citep{marinetraffic}}
    \label{tab:hercules_particulars}
    \begin{tabular}{ll}
    \hline
    IMO                    & 9642382      \\
    MMSI                   & 636015986    \\
    General Type           & Cargo        \\
    Specific Type          & Bulk Carrier \\
    Gross Tonnage          & 41,104       \\
    Summer Deadweight (t)  & 75,200       \\
    Length overall LOA (m) & 225          \\
    Width (m)              & 32.26        \\
    Beam (m)               & 32           \\
    Year Built             & 2013        
    \end{tabular}
\end{table}

%% file: hercules_power_specifications.tex
\begin{table}[h!]
    \centering
    \caption{Hercules Power Specifications}
    \label{tab:hercules_power_specifications}
    \begin{tabular}{ll}
    \hline
    Engine Type       & Diesel      \\
    Engine Builder    & MAN B\&W    \\
    Model             & 5S60MC-C    \\
    RPM               & 105         \\
    Stroke Type       & 2           \\
    Cylinder Stroke   & 2400        \\
    Cylinder Bore     & 600         \\
    Total Power (kw)  & 8833        \\
    Propulsion type   & Fixed Pitch \\
    Propulsion number & 1           \\
    Speed             & 14.5       
    \end{tabular}
    \end{table}

%% file: global_ocean_physics_analysis_and_forecast_data_description.tex
\begin{table}[!ht]
    \centering
    \caption{Global Ocean Physics Analysis and Forecast Data Description}
    \resizebox{\textwidth}{!}{%
    \begin{tabular}{ll}
    \hline
        \textbf{Dataset ID} & \textbf{Short Description} \\ \hline
        cmems\_mod\_glo\_phy\_anfc\_0.083deg\_P1D-m & 2D daily mean of various different measurements \\ 
        cmems\_mod\_glo\_phy-cur\_anfc\_0.083deg\_P1D-m & 3D daily mean horizontal currents information from top to bottom \\ 
        cmems\_mod\_glo\_phy-thetao\_anfc\_0.083deg\_P1D-m & 3D daily mean potential temperature information from top to bottom \\ 
        cmems\_mod\_glo\_phy-so\_anfc\_0.083deg\_P1D-m & 3D daily mean salinity information from top to bottom \\ 
        cmems\_mod\_glo\_phy-wcur\_anfc\_0.083deg\_P1D-m & 3D daily mean vertical current information from top to bottom \\ \hline
    \end{tabular}%
    }
    \label{tab:global_ocean_physics_analysis_and_forecast_data_description}
\end{table}

%% file: global_ocean_physics_analysis_and_forecast_raw_samples.tex
\begin{table}[!ht]
    \centering
    \caption{Global Ocean Physics Analysis and Forecast Raw Samples}
    \resizebox{\textwidth}{!}{%
    \begin{tabular}{llllrrrrr}
        \toprule
                &       &            &            &  \textbf{ist} &     \textbf{mlotst} &          \textbf{pbo} &  \textbf{siage} &  \textbf{sialb} \\
                \textbf{depth} & \textbf{latitude} & \textbf{longitude} & \textbf{time} &      &            &              &        &        \\
        \midrule
        0.494025 & -36.0 & -43.000000 & 2021-11-16 &  0.0 &  20.942480 &  5003.330078 &    0.0 &  0.066 \\
                &       & -42.916656 & 2021-11-16 &  0.0 &  20.706240 &  5009.921875 &    0.0 &  0.066 \\
                &       & -42.833328 & 2021-11-16 &  0.0 &  20.484440 &  5001.324219 &    0.0 &  0.066 \\
                &       & -42.750000 & 2021-11-16 &  0.0 &  19.690535 &  5000.277344 &    0.0 &  0.066 \\
                &       & -42.666656 & 2021-11-16 &  0.0 &  19.470037 &  5000.562988 &    0.0 &  0.066 \\
        \bottomrule
        \end{tabular}%
    }
    \label{tab:global_ocean_physics_analysis_and_forecast_raw_samples}
\end{table}

%% file: era5_hourly_on_single_levels_data_description.tex
\begin{table*}[!ht]
    \centering
    \caption{ERA5 Hourly on Single Levels Data Description}
    \resizebox{\textwidth}{!}{%
    \begin{tabular}{ll}
    \hline
        Data type & Gridded \\
        Projection & Regular latitude-longitude grid \\ 
        Horizontal coverage & Global \\ 
        Horizontal resolution & Reanalysis: 0.25° x 0.25° (atmosphere), 0.5° x 0.5° (ocean waves).  \\ 
        Temporal coverage & 1940 to present \\ 
        Temporal resolution & Hourly \\ 
        File format & GRIB \\ 
        Update frequency & Daily \\ \hline
    \end{tabular}%
    }
    \label{tab:era5_hourly_on_single_levels_data_description}
\end{table*}

%% file: era5_hourly_on_single_levels_raw_samples.tex
\begin{table}[!ht]
        \centering
        \caption{ERA5 Hourly on Single Levels Raw Samples}
        \resizebox{\textwidth}{!}{%
        \begin{tabular}{lllrrrrr}
        \toprule
                &       &            &  \textbf{number} &   \textbf{step} &  \textbf{surface} &      \textbf{u100} &       \textbf{v100} \\
                \textbf{latitude} & \textbf{longitude} & \textbf{time} &         &        &          &           &            \\
        \midrule
        56.0 & -43.0 & 2021-11-01 &       0 & 0 days &      0.0 &  4.457072 &   3.199743 \\
                &       & 2021-11-02 &       0 & 0 days &      0.0 &  4.228130 &  12.664192 \\
                &       & 2021-11-03 &       0 & 0 days &      0.0 &  6.418368 &   6.450352 \\
                &       & 2021-11-04 &       0 & 0 days &      0.0 &  4.549166 &   0.413618 \\
                &       & 2021-11-05 &       0 & 0 days &      0.0 &  3.050886 &  -1.523218 \\
        \bottomrule
        \end{tabular}%
        }
        \label{tab:era5_hourly_on_single_levels_raw_samples}
\end{table}

%% file: results.tex
\section{Results}\label{sec:results}

The comparative analysis revealed that \ac{rpm} outperformed \ac{sog} 
as a predictor of fuel consumption in the baseline model. 
This finding is supported both empirically (as shown by higher R² values and lower prediction errors) 
and theoretically by the fundamental mechanics of marine propulsion systems.
Specifically, \ac{rpm} is directly linked to the engine's rotational workload and thus 
to power output and fuel consumption, which often follows a cubic or higher-order nonlinear relationship 
with shaft speed \cite{alireza2018fuelengine}. 
In contrast, \ac{sog} is a kinematic parameter influenced by environmental forces such as wind, currents, 
and tides, which decouple it from the actual engine load and energy usage.

Linear and simpler models tend to outperform complex and non-linear ones. This is demonstrated in \cref{tab:baseline_performance_comparison_with_five_fold_cross_validation}. The table presents the baseline performance metrics of four regression models—Ridge Regression, Support Vector Regression (SVR), Random Forest, and XGBoost—evaluated using five-fold cross-validation. The comparison includes the mean and standard deviation of R², adjusted R², RMSE (Root Mean Squared Error), and MAE (Mean Absolute Error), offering a robust assessment of both model accuracy and consistency. Among the evaluated models, Ridge Regression outperforms the others, achieving the highest mean R² (0.7606) and adjusted R² (0.7597) values, indicating a stronger explanatory power with minimal overfitting. It also attains the lowest average RMSE (5.9762) and MAE (3.1582), suggesting superior predictive accuracy and lower average error in fuel consumption estimation. SVR ranks second across most metrics, with a slightly lower R² (0.7261) and adjusted R² (0.7251), along with a higher RMSE (6.4549) and MAE (3.3779), indicating moderately reduced performance relative to Ridge Regression. Random Forest performs worse than both Ridge and SVR, particularly in terms of error metrics, while XGBoost shows the lowest performance overall, with the smallest R² (0.6062) and highest RMSE (7.6537) and MAE (4.1084), indicating less effective generalization to the data. In terms of variability, all models demonstrate comparable standard deviations across folds, with Ridge Regression maintaining the lowest variance in both RMSE and MAE, underscoring its reliability and stability.

Feature analysis revealed that adding environment variables proved beneficial in predicting FC compared to baseline as shown in \cref{tab:advanced_model_performance_comparison_with_five_fold_cross_validation}. Among the evaluated models, Random Forest yields the best overall performance, achieving the highest mean R² (0.9021) and adjusted R² (0.9006), indicating superior explanatory power. It also records the lowest RMSE (3.5385) and MAE (2.1864), underscoring its high predictive accuracy and minimal average deviation between predicted and actual fuel consumption values. Additionally, the relatively low standard deviations across folds highlight the model’s consistency. Ridge Regression, XGBoost, and SVR exhibit comparable predictive performance, with R² values ranging from 0.8906 to 0.8941 and similar RMSE values (~3.71 for Ridge and XGBoost; 3.64 for SVR). However, SVR attains a slightly higher adjusted R² (0.8925) than Ridge and XGBoost and shows a lower MAE than both, suggesting that SVR may yield marginally better performance in minimizing absolute prediction errors with slightly greater variability across folds (MAE std: 0.4149). Overall, the results underscore that ensemble-based methods like Random Forest are particularly effective in modeling fuel consumption when more advanced modeling strategies are employed. The transition from baseline to advanced models led to a significant improvement in predictive performance across all metrics. In the baseline setup, Ridge Regression performed best, but in the advanced configuration, Random Forest emerged as the top performer with a notable increase in R² (from 0.76 to 0.90) and substantial reductions in RMSE and MAE. These improvements reflect enhanced model accuracy, better generalization, and increased stability.

\begin{figure}[htbp]
    \centering
    \includegraphics[width=\linewidth]{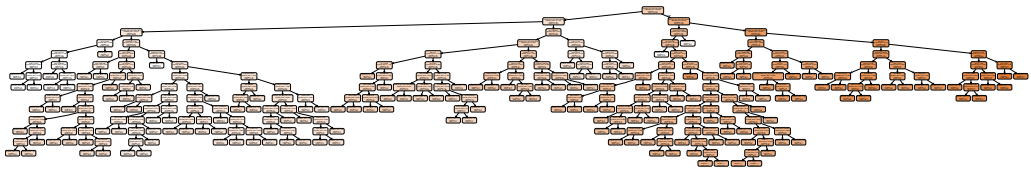}
    \caption{Random Forest Tree}
    \label{fig:random_forest_tree}
\end{figure}

\begin{figure}[htbp]
    \centering
        \includegraphics[width=1\textwidth]{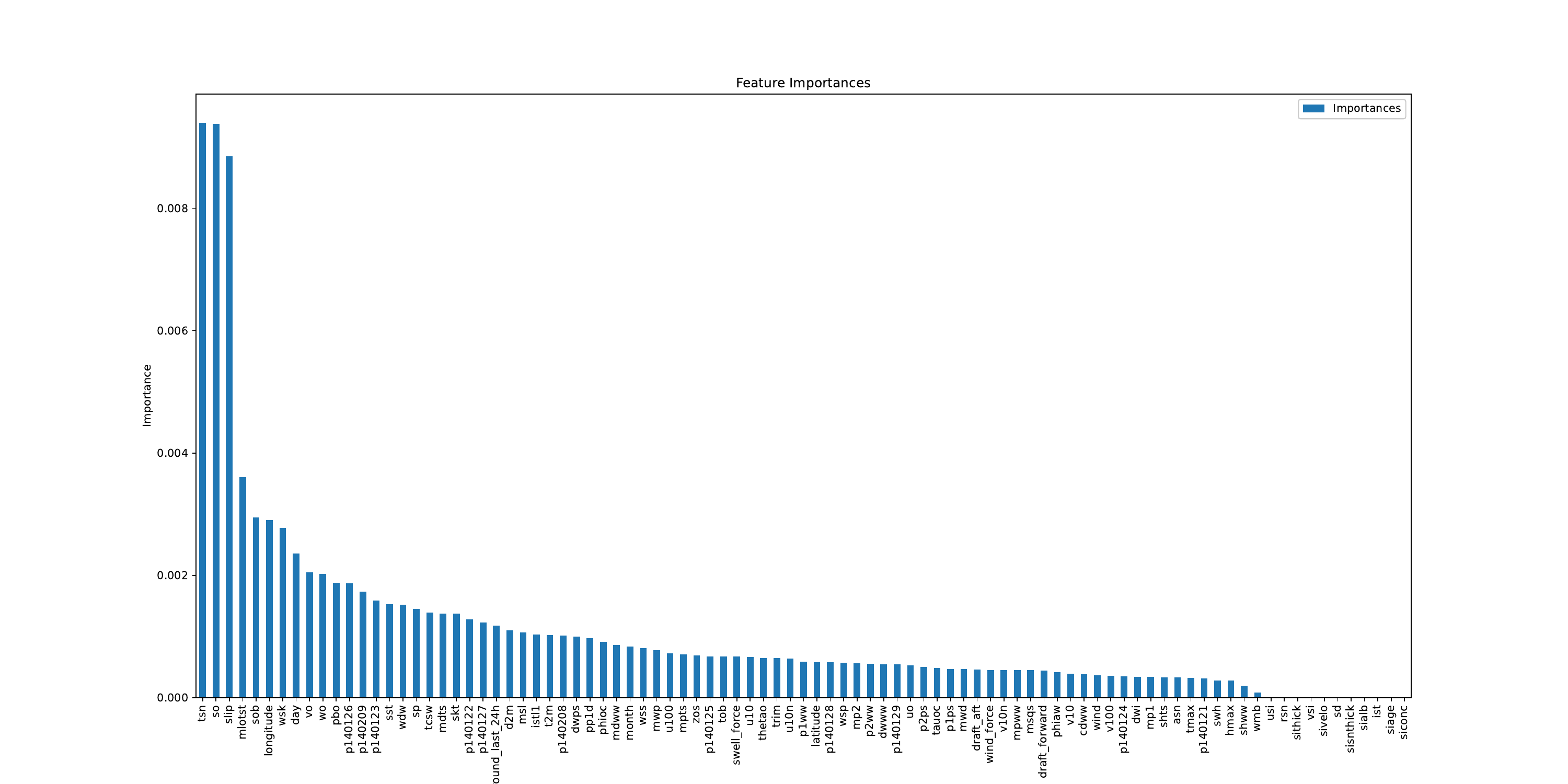}
        \caption[Features Importance]{Excluding Engine RPM}
        \label{fig:features_importance}
\end{figure}

Most influencial parameters that contributed significantly to the overall accuracy and were used in building the advanced model in the order of importance are Engine RPM, Temperature in surface snow, seawater salinity, and Propeller slip. All other features contributed somewhat but needed to be more significant to justify their usage in the final model, and therefore, they were not used. These findings are depicted in \cref{fig:random_forest_tree,fig:features_importance}. The bar plot presents the feature importance scores derived from a tree-based model Random Forest, used for predicting fuel consumption. The feature importance metric reflects the contribution of each input variable to reducing the prediction error, and hence to the overall model performance. In order to not skew the figure the Engine RPM have been filtered out from the view. The most influential features are "tsn", "tsr", and "spd", each exhibiting the highest importance scores and their dominant contribution emphasizes the critical role of environment-related metrics in determining fuel consumption. Following these are meteorological and geospatial variables such as "mssl", "mstp", and "longitude", along with temporal features like "day" and "wo" (week of observation). This indicates that, beyond machinery metrics, environmental conditions and voyage context also play a non-negligible role in shaping fuel use dynamics. Several oceanographic and weather-related parameters (e.g., "p140208", "ppbd", "mwd", "mwsh") fall into a mid-importance range, suggesting their secondary influence on consumption, potentially through effects on vessel resistance or route adjustments. Conversely, the features toward the right end of the plot (e.g., "sicom", "sige", "ssmtbkc") exhibit minimal importance, implying limited predictive value in the current modeling setup.

\begin{figure}[htbp]
    \centering
        \includegraphics[width=1\textwidth]{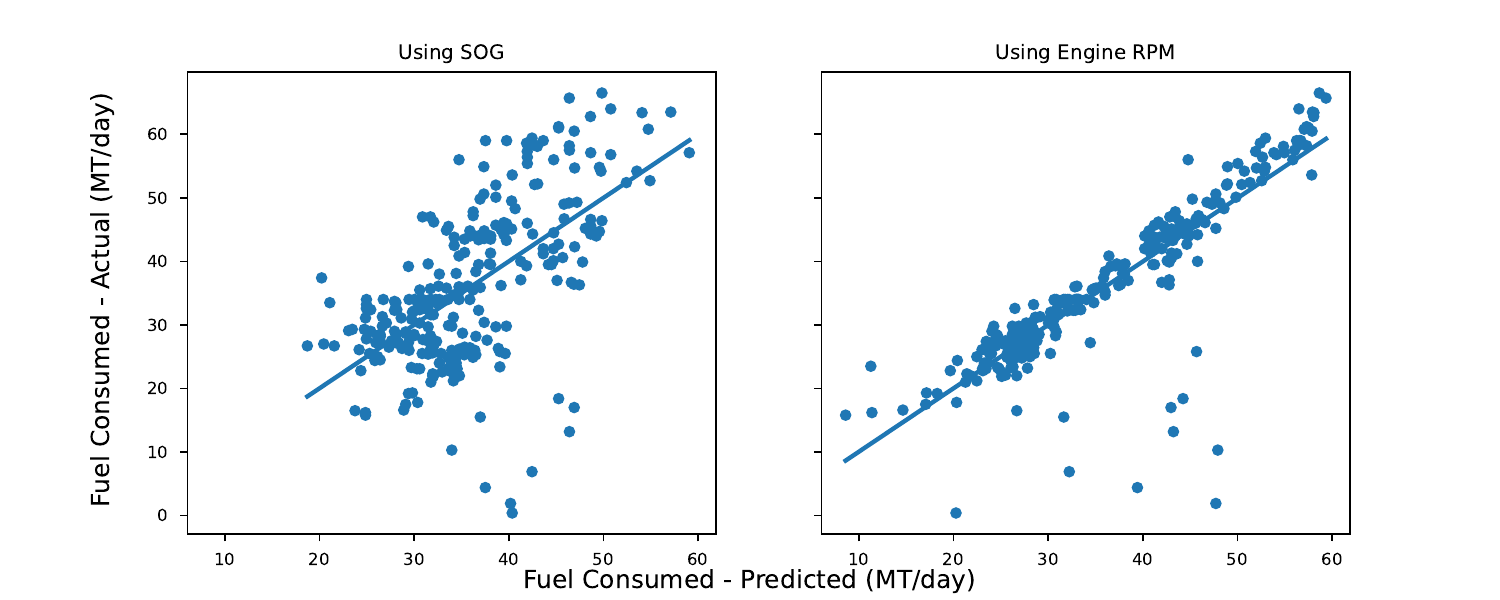}
        \caption[Baseline Ridge to Compare Fit]{SOG vs Engine RPM}
        \label{fig:baseline_ridge_to_compare_fit_using_sog_vs_engine_rpm}
\end{figure}

\input{baseline_performance_comparison_with_five_fold_cross_validation}
\input{advanced_model_performance_comparison_with_five_fold_cross_validation}

\begin{figure}
\begin{lstlisting}[
        language=Python,
        firstline=13,
        lastline=19,
        caption={Grid Seach Hyperparameters Space},
        label={src:grid_search_hyperparameters_space}
    ]
HYPER_PARAMETERS = {
    'rf__n_estimators': [50, 100, 200],       
    'rf__max_depth': [None, 10, 20, 30],      
    'rf__min_samples_split': [2, 5, 10],      
    'rf__min_samples_leaf': [1, 2, 4],        
    'rf__bootstrap': [True, False]            
}
\end{lstlisting}
\end{figure}

%% file: baseline_performance_comparison_with_five_fold_cross_validation.tex
\begin{table}[!ht]
        \centering
        \caption{Baseline Performance Comparison with Five Fold Cross}
        \resizebox{\textwidth}{!}{%
        \begin{tabular}{lrrrrrrrr}
        \toprule
        {} &   \textbf{r2\_mean} &    \textbf{r2\_std} &  \textbf{adj\_r2\_mean} &  \textbf{adj\_r2\_std} &  \textbf{rmse\_mean} &  \textbf{rmse\_std} &  \textbf{mae\_mean} &   \textbf{mae\_std} \\
        \midrule
        ridge        &  0.760612 &  0.130927 &     0.759735 &    0.131407 &   5.976210 &  2.002627 &  3.158223 &  0.734065 \\
        svr          &  0.726106 &  0.126659 &     0.725103 &    0.127123 &   6.454855 &  1.918580 &  3.377862 &  0.832788 \\
        randomforest &  0.690246 &  0.138696 &     0.689111 &    0.139204 &   6.792727 &  2.029012 &  3.710294 &  0.712134 \\
        xgboost      &  0.606152 &  0.162200 &     0.604709 &    0.162794 &   7.653745 &  2.014358 &  4.108379 &  0.774200 \\
        \bottomrule
        \end{tabular}%
        }
        \label{tab:baseline_performance_comparison_with_five_fold_cross_validation}
\end{table}

%% file: advanced_model_performance_comparison_with_five_fold_cross_validation.tex
\begin{table}[!ht]
        \centering
        \caption{Advanced Model Performance Comparison with Five Fold Cross}
        \resizebox{\textwidth}{!}{%
        \begin{tabular}{lrrrrrrrr}
        \toprule
        {} &   \textbf{r2\_mean} &    \textbf{r2\_std} &  \textbf{adj\_r2\_mean} &  \textbf{adj\_r2\_std} &  \textbf{rmse\_mean} &  \textbf{rmse\_std} &  \textbf{mae\_mean} &   \textbf{mae\_std} \\
        \midrule
        randomforest &  0.902077 &  0.057732 &     0.900576 &    0.058617 &   3.538472 &  1.140355 &  2.186424 &  0.231431 \\
        ridge        &  0.891360 &  0.064465 &     0.889695 &    0.065453 &   3.716946 &  1.182466 &  2.355040 &  0.359092 \\
        xgboost      &  0.890620 &  0.063637 &     0.888944 &    0.064612 &   3.716129 &  1.194261 &  2.352869 &  0.472801 \\
        svr          & 0.894117	 &  0.069662 &    0.892494  &    0.070730 &  3.639306 &  1.263992 &  2.251402  &  0.414863 \\
        \bottomrule
        \end{tabular}%
        }
        \label{tab:advanced_model_performance_comparison_with_five_fold_cross_validation}
\end{table}

%% file: conclusion.tex
\section{Conclusion}\label{sec:conclusion}

This study confirmed the massive potential of using \ac{ml} methods and fusing external data to model Ship Fuel Consumption, which is the basis for all ship optimization approaches. Additional environmental factors also proved helpful in improving the predictive power. However, only a small subset of factors had a significant impact, with internal parameters such as engine \ac{rpm} being the most predictive factor for \ac{fc}. 
Most of the features have not contributed to the overall predictive power, which is good that with fewer features, the \ac{fc} could be, to some extent, accurately predicted. Still, on the other hand, this raises the question of more investigation regarding the extensive data preparation explicitly required on the environmental parameters, for example, computing ship heading and then the angle of attack relative to the vessel. Moreover, more voyage report samples are needed to drill down on those features. Additionally, the principles developed in this study could be tested against different vessels of the same class to validate generalization on a broader scope. 
Additionally other marine fuels should be explored. Currently, the target ships of \ac{fc} models are mostly diesel-powered, and the fuel consumption is primarily diesel. With the emergence of new or hybrid energy ships, more energy-efficient fuels such as LNG, fuel cells, methanol, and ammonia can be adopted to save ships energy.
The \ac{fc} models should be combined with energy efficiency optimization methods. The ship energy efficiency optimization method can be divided into the supply side (inside the ship) and the demand side (outside the ship), corresponding to the energy management and operation optimization strategies. Thus it would be interesting to explore some data sources of bunker fuel consumption with a finer data granularity, such as sensor data, and the benefits of combining these data sources with other data sources that provide complementary information.

An advanced model is developed using the features identified by the \ac{rf}. Two significant insights have been derived. First, non-linear models tend to perform much better than baseline models. Secondly, the performance significantly rose from 76 percent using the baseline model to 90 percent, 
which highlights that with a handful of features, 90 percent of the \ac{fc} could be accurately estimated.

The comparison clearly indicates that advanced modeling approaches result in superior predictive accuracy, lower error metrics, and more stable performance across validation folds. The transition from baseline to advanced models, particularly using more informative variables plays a critical role in enhancing fuel consumption estimation models. The findings reinforce the importance of model selection, feature engineering, and hyperparameter tuning in achieving reliable maritime fuel prediction.

Nevertheless, all models demonstrate strong predictive capability, indicating the effectiveness of machine learning approaches for this application.

Overall, the results demonstrate that fuel consumption prediction is most sensitive to 
engine and propulsion parameters, with contextual and environmental 
variables providing complementary but lesser explanatory power. 
These insights can guide feature selection and data acquisition strategies in future model refinement.

Another angle would be to investigate how sailing in unpopular regions like the Arctic could influence \ac{fc} since it has been observed in this study that seawater temperature and salinity played massive role in the \ac{fc} and ultimately to act as a guide to suggest a most efficient route for maritime navigation. 

To streamline further research on the ideas developed in this study, a modular and comprehensive package has been developed that provides a baseline to build on top of for reproducible research and additional investigations.   

This study has been an attempt to contribute toward environmentally sustainable shipping operations. 

%% file: voyage_reports_raw_parameters_and_unit_of_measure.tex
\begin{table}[!ht]
    \centering
    \caption{Voyage Reports Raw Parameters and Unit of Measure}
    \resizebox{\textwidth}{!}{%
    \begin{tabular}{ll}
        \toprule
        \textbf{Parameter}                                        & \textbf{Unit of Measurement}        \\ 
    \midrule
    Date/time (08:00/16:00 Geneva time)              & Daily                      \\
    Geograph. position                               & Degree and Decimal Minutes \\
    Current Speed                                    & knots                      \\
    Average speed 24hrs                              & knots                      \\
    Average speed since departure                    & knots                      \\
    Average RPM last 24hrs                           & knots                      \\
    Propeller Slip                                   & Percentage                 \\
    Wind Direction                                   & Compass Points             \\
    Wind Force                                       & Beaufort Scale             \\
    Swell Direction                                  & Cardinal Direction         \\
    Swell Force                                      & Beaufort Scale             \\
    Current Direction                                & Cardinal Direction         \\
    Current Speed                                    & knots                      \\
    ULSFO bunkers consumption last 24hrs - Main Engine & MT (metric tone)         \\
    ULSFO bunkers consumption last 24hrs - Boiler    & MT                         \\
    MGO bunkers consumption last 24hrs - Main Engine & MT                         \\
    MGO bunkers consumption last 24hrs - Boiler      & MT                         \\
    MGO bunkers consumption last 24hrs - Auxiliary   & MT                         \\
    Bunkers ROB ULSFO \textless 0.5\%Sulfur          & MT                         \\
    Bunkers VLSFO  \textless 0.5\%Sulfur             & MT                         \\
    Bunkers ROB LSMGO \textless 0.1\%Sulfur          & MT                         \\
    FW ROB                                           & MT                         \\
    Next port DTG                                    & Time                       \\
    Next port ETA                                    & Time                       \\
    Draft Current FWD/AFT                            & Ratio                      \\
    Draft Arrival FWD/AFT                            & Ratio                      \\
    Next port Name                                   & String                     \\
    Current Adverse / Favorable                      & String                     \\
    Suppliers required in next port                  & String                     \\
    \bottomrule
    \end{tabular}%
    }
    \label{tab:voyage_reports_raw_parameters_and_unit_of_measure}
    \end{table}

%% file: cmems_raw_parameters_and_unit_of_measure.tex
\begin{table}[!ht]
    \centering
    \caption{CMEMS Raw Parameters and Unit of Measure}
    \label{tab:cmems_raw_parameters_and_unit_of_measure}
    \resizebox{\textwidth}{!}{%
    \begin{tabular}{lll}
    \toprule
    \textbf{Parameter} & \textbf{Description}                                        & \textbf{Unit of Measurement}    \\
    \midrule 
    ist       & Sea ice surface temperature                        & °C                     \\
    mlotst    & Ocean mixed layer thickness defined by sigma theta & m                      \\
    pbo       & Sea water pressure at sea floor                    & dbar                   \\
    siage     & Age of sea ice                                     & year                   \\
    sialb     & Sea ice albedo                                     & \%                     \\
    siconc    & Sea ice area fraction                              & ratio                  \\
    sisnthick & Surface snow thickness                             & m                      \\
    sithick   & Sea ice thickness                                  & m                      \\
    sivelo    & Sea ice speed                                      & m/s                    \\
    sob       & Sea water salinity at sea floor                    & 10-3                   \\
    tob       & Sea water potential temperature at sea floor       & °C                     \\
    usi       & Eastward sea ice velocity                          & m/s                    \\
    vsi       & Northward sea ice velocity                         & m/s                    \\
    zos       & Sea surface height above geoid                     & m                      \\
    uo        & Eastward sea water velocity                        & m/s                    \\
    vo        & Northward sea water velocity                       & m/s                    \\
    so        & Sea water salinity                                 & 10\textasciicircum{}-3 \\
    thetao    & Sea water potential temperature                    & °C                     \\
    wo        & Upward sea water velocity                          & m/s                    \\
    \bottomrule
    \end{tabular}%
    }
    \end{table}

%% file: era5_raw_parameters_and_unit_of_measure.tex
\begin{table}[!ht]
    \centering
    \caption{ERA5 Raw Parameters and Unit of Measure}
    \label{tab:era5_raw_parameters_and_unit_of_measure}
    \resizebox{\textwidth}{!}{%
    \begin{tabular}{lll}
    \toprule
    \textbf{Parameter} & \textbf{Description} & \textbf{Unit of Measurement}    \\
    \midrule 
    tsn       & Temperature in surface snow & K                   \\
    \bottomrule
    \end{tabular}%
    }
    \end{table}
    